\documentclass[runningheads]{llncs}
\usepackage{graphicx}
\usepackage{float}
\usepackage{amsmath,amsfonts,amssymb}
\newcommand{\sign}{\text{sign}}
\usepackage{caption}
\begin{document}
%
\title{A Deep Genetic Programming based Methodology for Art Media Classification Robust to Adversarial Perturbations}
\titlerunning{A Deep GP Methodology Robust to Adversarial Perturbations}

%
\author{Gustavo Olague \inst{1} \and
Gerardo Ibarra-Vázquez\inst{2} \and
Mariana Chan-Ley \inst{1} \and
Cesar Puente\inst{2} \and
Carlos Soubervielle-Montalvo\inst{2} \and
Axel Martinez \inst{1}
}

\institute{EvoVisión Laboratory, CICESE Research Center. Carretera Ensenada-Tijuana 3918, Zona Playitas, 22860, Ensenada, B.C., México
\and
Universidad Autónoma de San Luis Potosí, Facultad de Ingeniería. Dr. Manuel Nava 8, Col. Zona Universitaria Poniente, 78290, San Luis Potosí, S.L.P., México}
\authorrunning{G. Olague et al.}
%
%
\maketitle              
\begin{abstract}
Art Media Classification problem is a current research area that has attracted attention due to the complex extraction and analysis of features of high-value art pieces. The perception of the attributes can not be subjective, as humans sometimes follow a biased interpretation of artworks while ensuring automated observation's trustworthiness. Machine Learning has outperformed many areas through its learning process of artificial feature extraction from images instead of designing handcrafted feature detectors. However, a major concern related to its reliability has brought attention because, with small perturbations made intentionally in the input image (adversarial attack), its prediction can be completely changed. In this manner, we foresee two ways of approaching the situation: (1) solve the problem of adversarial attacks in current neural networks methodologies, or (2) propose a different approach that can challenge deep learning without the effects of adversarial attacks. The first one has not been solved yet, and adversarial attacks have become even more complex to defend. Therefore, this work presents a Deep Genetic Programming method, called Brain Programming, that competes with deep learning and studies the transferability of adversarial attacks using two artworks databases made by art experts. The results show that the Brain Programming method preserves its performance in comparison with AlexNet, making it robust to these perturbations and competing to the performance of Deep Learning.

\end{abstract}
\section{Introduction}

Art media refers to the materials and techniques used by an artist to create an artwork. The categorization problem of visual art media is an open research area with challenging tasks, such as the classification of fine art pieces, which is extremely difficult due to the selection of features that distinguish each medium. For example, an art expert analyzes the style, genre, and media from artworks to classify them.

The artwork style is associated with the author's school and is usually described by its distinctive visual elements, techniques, and methods. Recognition of the form is related to the localization of features at different levels. The classical hierarchy of genres ranks history-painting and portrait as high, while landscapes and still-life are low because they did not contain persons. Therefore, handling these many aspects of an automated classification system is a big challenge.    

The recent progress of Machine Learning (ML) in Computer Vision (CV) tasks has made methodologies such as Deep Learning (DL) adaptable to many research areas like the categorization problem of art media. Commonly, these methodologies learn from the visual content and contextual information of the image to assign the class or category to which it belongs. DL is known to achieve exemplary performance in many areas.  However, recent studies have demonstrated that Adversarial Attacks (AA) pose a predicting threat to DL's success because with small perturbations intentionally created, they could lead to incorrect outputs to a model. 

In this matter, AA is a popular research topic covering all aspects of the attack architectures and defense mechanisms to diminish the attack damage. Nevertheless, despite significant efforts to solve this problem, attacks have become more complex and challenging to defend. Today, researchers study AA from different viewpoints. On the one hand, white-box attacks refer to when the targeted model is known, including its parameter values, architecture, and training method. On the other hand, black-box attacks are when AA generates adversarial examples or perturbed images with no information on the targeted architecture model during learning \cite{akhtar2018threat}. Another feature of the attacks is that it can be specifically designed to predict a desirable class (targeted attack) or produce an incorrect output no matter the class (untargeted attack). Furthermore, it has been reported that AA can be transferable from an ML model to others. Hence, we foresee two ways to approach the situation: (1) solve the problem of adversarial attacks in current neural networks methodologies, or (2) propose a different approach that can challenge deep learning by being immune to adversarial attacks.

This article presents a study of the transferability and the effects of adversarial attacks made for deep learning to an approach that solves the problem of image classification through a genetic programming based (GP-like) methodology called ``Brain Programming" (BP) (explained in Section~\ref{Methodology}). Extend the study of the effects of \textit{adversarial attacks} on a different approach for image classification would highlight the differences between performance and robustness to these perturbations.

\section{Related Research}

The categorization problem of art media in CV has arisen from the need to have automatic systems for identifying valuable artwork pieces to have a trustworthy analysis that can not be subjective as humans are prone to be. Firstly, handcrafted feature extraction approaches were used to solve the problem. One of the first approaches \cite{keren2002painter} proposed a Discrete Cosine Transform (DCT) coefficients scheme for feature extraction for painter identification by classifying the artist's style. They were able to find five painters among 30 artworks with encouraging but not perfect results. 

Later, wavelets were used to analyze several features from artworks like texture, geometry, style, brush strokes, and contours. In \cite{li2004studying}, artist classification was made using wavelets of brush strokes drawn on ancient paintings. In \cite{johnson2008image}, wavelets were used with several classification algorithms such as support vector machines (SVM), hidden Markov models, among others for artist identification of 101 high-resolution grayscale paintings. In \cite{arora2012towards}, it is presented a comparative study of different classification methodologies based handcrafted features such as semantic-level features with an SVM, color SIFT (Scale-Invariant Feature Transform) and opponent SIFT with bag-of-words and latent Dirichlet allocation with a generative bag-of-words topic model for fine-art genre classification.

Recently, ML approaches benefit from the learning process to acquire features from the images' visual content. For example, in \cite{kowaliw2010evolutionary}, they proposed a GP method that uses transform-based evolvable features to evolve features that are evaluated through a standard classifier. In \cite{ChanLey2020CategorizationOD,Olague2020}, authors reported using a GP-like methodology that aims to emulate the behavior of the brain based on neuroscience knowledge for art media categorization having competitive results with a DL model. Nevertheless, approaches based on convolutional neural network (CNN) became famous because of their outstanding performances in many areas. Bar et al\cite{bar2014classification} proposed a compact binary representation combined with the PiCoDes descriptors from a deep neural network to identify artistic styles in paintings showing exceptional results on a large-scale collection of paintings.  In \cite{yang2020classification}, they employ a deep CNN to recognize artistic media from artworks and classify them into several categories such as oil-paint brush, pastel, pencil, and watercolor. They compare their results with that of trained humans having comparable results.

Thus, even CNN architectures have classified large-scale sets of images with multiple classes with similar results that trained humans, the security concerns about these architectures make them unreliable. The brittleness is because, with small perturbations produced on the image, DL can be intentionally fooled. For example, there are critical areas in museums and galleries such as artist identification and forgery detection, where the confidence of the prediction must not depend on a system that can be manipulated by an imperceptible perturbation. This catastrophic scenario could lead to forgeries to circulate on the market or be misattributed to a specific artist.

\section{Problem Statement}\label{problemstatement}

In this section, we detail the serious problem in the DL structure to the adversarial attacks. First, given an input image $\mathbf{x}$ and its corresponding label $y$, DL establish a relationship within the data by the following equation:

\begin{equation}
    y=f(\mathbf{x})=\mathbf{w}^\intercal \mathbf{x} \;\;\;,
\end{equation}

where function $f()$ is the DL model, whose associated weights parameters are $\mathbf{w}$. However, an erroneous behavior is notable when the input image suffers a small change in its pixels $\mathbf{x}_{\mathbf{\rho}} = \mathbf{x} + \rho$ such that:

\begin{align}
    f(\mathbf{x}) \neq f(\mathbf{x}_{\mathbf{\rho}}) \\
    ||\mathbf{x} - \mathbf{x}_{\mathbf{\rho}}||_{p}<\alpha
\end{align}

where $p\in N ,p\geq 1, \alpha \in R, \alpha\geq 0$. So, it can be defined an Adversarial Example as an intentional modified input $\mathbf{x}_{\mathbf{\rho}}$ that is classified differently than $\mathbf{x}$ by the DCNN model, with a limited level of change in the pixels of $||\mathbf{x} - \mathbf{x}_{\mathbf{\rho}}||_{p}<\alpha$, so that it may be imperceptible to a human eye. 

The simplest explanation of how adversarial examples works to attack DL models is that most digital images use 8‑bit per channel per pixel. So, each step of 1/255 limits the data representation; the information in between is not used. Therefore, if every element of a perturbation $\rho$ is smaller than the data resolution, it is coherent for the linear model to predict distinct given an input $\mathbf{x}$ than to an adversarial input $\mathbf{x}_{\mathbf{\rho}} = \mathbf{x} + \rho$. We assume that forasmuch as $||\mathbf{\rho}||_{\infty} <\alpha$, where $\alpha$ is too small to be discarded, the classifiers should predict the same class to $\mathbf{x}$ and $\mathbf{x}_{\mathbf{\rho}}$.

Nonetheless, after applying the weight matrix $\mathbf{w} \in \mathbf{R}^{M \times N}$ to the adversarial example, we obtain the dot product defined by $\mathbf{w}^\intercal \mathbf{x}_{\mathbf{\rho}}= \mathbf{w}^\intercal\mathbf{x} + \mathbf{w}^\intercal\mathbf{\rho}$. Hence, the adversarial example will grow the activation by $\mathbf{w}^\intercal\mathbf{\rho}$. Note that the dimensionality of the problem does not grow with $||\mathbf{\rho}||_{\infty}$; thus, the activation change caused by perturbation $\mathbf{\rho}$ can grow linearly with $n$. As a result, the perturbation can make many imperceptible changes to the input to obtain big output changes.

DL's behavior is hugely linear to be immune to adversarial examples, and nonlinear models such as sigmoid networks are set up to be in the non-saturating most of the time, becoming them more like a linear model. Hence, every perturbation as accessible or challenging to compute should also affect deep neural networks. Therefore, when a model is affected by an adversarial example, this image often affects another model, no matter if the two models have different architectures or were trained with different databases. They just have to be set up for the same task to change the result \cite{Goodfellow2015ExplainingAH}.

\section{Methodology}\label{Methodology}

The experiment consists of studying the transferability of an AA from CNN to BP. This problem's methodology considers unconventional training, validation, and test databases since we apply two different image databases compiled by art experts. Training and validation databases are constructed from the Kaggle database, while testing uses a standard database WikiArt (See Table~\ref{Tab:dataset}). The aim is to emulate a real-world scenario where the trained models are tested with an unseen standard benchmark compiled by a different group of experts.

Validation and test databases are used to compute adversarial examples using the fast gradient signed method (FGSM) and AlexNet architecture using standard values for scale $\epsilon={2,4,8,16,32}$ to build the perturbations. The implementation of the FGSM was made on Pytorch v1.1. AlexNet was trained using transfer learning with the pre-trained model from Pytorch, and BP utilizes the models reported in \cite{ChanLey2020CategorizationOD,Olague2020}.

We formulate the art media classification problem in terms of a binary classification, whose main goal is to find the class elements. Also, we employ classification accuracy as a measure of performance for the classifiers, which is simply the rate of correct classifications given by the following formula:

\begin{equation*}
    Accuracy= \frac{1}{N}\sum_{n=1}^{N} d(y\prime_{n},y_{n})
\end{equation*}

\noindent
where $N$ is the total of test images, $y\prime_{n}$ is the predicted label for the image $n$, $y_{n}$ is the original label for the image $n$, and $d(x,y)=1$ if $x=y$ and 0 otherwise. In the following sections, the methods used for the experiment are briefly explained.

\subsection{Brain Programming}\label{bp}

BP is an evolutionary paradigm for solving CV problems that is reported in \cite{HERNANDEZ2016216,Olague2019,Olague2019b}. This methodology extracts characteristics from images through a hierarchical structure inspired by the brain's functioning.  BP proposes a GP-like method, using a multi-tree representation for individuals. The main goal is to obtain a set of evolutionary visual operators ($EVOs$), also called visual operators ($VOs$), which are embedded within a hierarchical structure called the artificial visual cortex (AVC). 

\begin{figure}
\includegraphics[width=\textwidth]{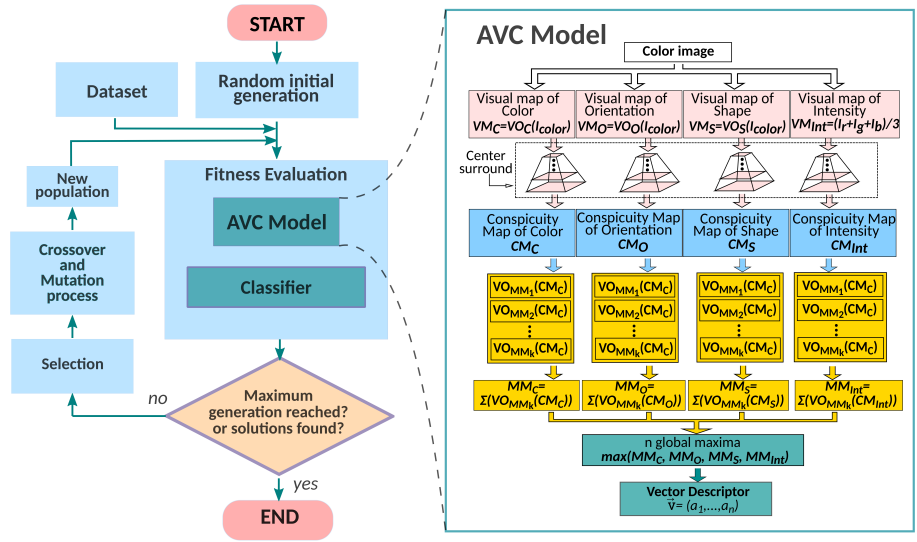}
\caption{Brain Programming workflow}
\label{fig:mapsmodel}
\end{figure}

BP can be summarized in two steps: first, the evolutionary process whose primary purpose is to discover functions to optimize complex models by adjusting the operations within them. Second, the AVC, a hierarchical structure inspired by the human visual cortex, uses the concept of composition of functions to extract features from images. The model can be adapted depending on the task, whether it is trying to solve the focus of attention for saliency problems or the complete AVC for categorization/classification problems. In this section, we briefly described the BP workflow (see Figure \ref{fig:mapsmodel}), but further details are explained in \cite{ChanLey2020CategorizationOD}.

\subsubsection{Initialization}

First, we set the parameters of the evolutionary process of BP and establish the image databases. Next, a random initial population is created to evolve the population. In BP, an individual is a computer program represented by syntactic trees embedded into a hierarchical structure.

Individuals within the population contain a variable number of syntactic trees, ranging from 4 to 12, one for each evolutionary visual operator ($VO_O$, $VO_C$, $VO_S$, $VO_I$) regarding orientation, color, shape, and intensity; and at least one tree to merge the visual maps produced and generate the Mental Maps (MM). Details about the usage of these visual operators are explained in detail in \cite{HERNANDEZ2016216,Olague2019,ChanLey2020CategorizationOD}.

Functions within each $VO$ are defined with expert knowledge to attend characteristics related to the dimension they represent and updated through genetic operations. After creating the first generation, the AVC model is used to evaluate the population's fitness, as shown in Figure \ref{fig:mapsmodel}.

\subsubsection{Fitness function: Artificial Virtual Cortex (AVC)}
\label{sec:AVC}

The evolutionary loop starts evaluating each individual by using the $VOs$ generated in the previous step to extract features from input images through the AVC structure depicted in Figure \ref{fig:mapsmodel}. The result of this procedure is a descriptor vector that encodes the object. Then, BP uses an SVM to calculate the classification rate for a given training image database. We explain the detailed steps below. The entrance to the system is an RGB image that belongs to a predefined class. This system follows a function-based instead of data-based paradigm; hence, we define an image $ I $ as the graph-of-a-function. 

\textbf{Definition 1. Image as the graph of a function}. \textit{Let $f$ be a function $f:U \subset \mathbb{R}^2 \rightarrow \mathbb{R}$. The graph or image $I$ of $f$ is the subset of $\mathbb{R}^3$ that consist of the points $(x, y, f(x,y))$, in which the ordered pair $(x,y)$ is a point in $U$ and $f(x,y)$ is the value at that point. Symbolically, the image $I = \{(x,y,f(x,y)) \in \mathbb{R}^3 | (x,y) \in U\}$}.

This definition is based on the fact that the images result from the impression of variations in light intensity along the two-dimensional plane.

\subsubsection{Visual Maps}

Each input image is transformed to build the set $I_{color}$ = $\{I_r$, $I_g$, $I_b$, $I_c$, $I_m$, $I_y$, $I_k$, $I_h$, $I_s$, $I_v\}$, where each element corresponds to the color components of the RGB (red, green, blue), CMYK (Cyan, Magenta, Yellow, and black) and HSV (Hue, Saturation, and Value) color spaces. Elements on $I_{color}$ are the inputs to four $VOs$ defined by each individual. It is important to note that each solution in the population should be understood as a complete system and not only as a list of three-based programs. Individuals represent a possible configuration for feature extraction that describes input images and are optimized through the evolutionary process. Each $VO$ is a function applied to the input image to extract specific features from it, along with information streams of color, orientation, shape, and intensity; each of these properties is called a dimension. The output to $VO$ is an image called Visual Map ($VM$) for each dimension.

\subsubsection{Conspicuity Maps}

The next step is the center-surround process; it efficiently combines the information from the $VMs$ and is useful for detecting scale invariance in each of the dimensions. This process is performed by applying a Gaussian smoothing over the $VM$ at nine scales; this processing reduces the visual map's size by half on each level forming a pyramid. Subsequently, the six levels of the pyramid are extracted and combined. Since the levels have different sizes, each level is normalized and scaled to the visual map's dimension using polynomial interpolation. This technique simulates the center-surround process of the biological system. After extracting features, the brain receives stimuli from the vision center and compares it with the receptive field's surrounding information. The goal is to process the images so that the results are independent of scale changes. The entire process ensures that the image regions are responding to the indicated area. This process is carried out for each characteristic dimension; the results are called Conspicuity Maps ($CM$), focusing only on the searched object by highlighting the most salient features.

\subsubsection{Mental Maps}

Following the AVC flowchart, all information obtained is synthesized to build maps that discriminate against the unwanted information previously computed by the $CMs$. These new maps are called Mental Maps ($MMs$).

The AVC model uses a set-of-functions to extract the images' discriminant characteristics; it uses a functional approach. Thus, a set of $k$ $VOs$ is applied to the $CMs$ for the construction of the $MMs$. These $VOs$ correspond to the remaining part of the individual that has not been used. Unlike the operators used for the $VMs$, the operators' whole set is the same for all the dimensions. These operators filter the visual information and extract the information that characterizes the object of interest. Then, using Equation (\ref{eq:MM}), where $d$ is the dimension, and $k$ represents the cardinality of the set of $VO_{MM_k}$, we apply the $MMs$ for each dimension.

\begin{equation} \label{eq:MM}
MM_d = \sum_{i=1}^k VO_{MM_i}\left(CM_d\right)
\end{equation}

\subsubsection{Descriptor vector and classification}

The following stage in the model is the construction of the image descriptor vector ($DV$). The system concatenates the four $MMs$ and uses a max operation to extract the $n$ highest values; these values are used to construct the $DV$. Once we get the descriptor vectors of all the images in the database, the system trains an SVM. The classification score obtained by the SVM indicates the fitness of the individual.

\subsubsection{Selection and Reproduction}

A set of individuals is selected from the population with a probability based on fitness to participate in the genetic recombination, and the best individual is retained for further processing. The new individual of the population is created from the selected individual by applying genetic operators. Like genetic algorithms, BP executes the crossover between two selected parents at the chromosome level using a "cut-and-splice" crossover. Thus, all data beyond the selected crossover point is swapped between both parents A and B. 
The result of applying a crossover at the gene level is performed by randomly selecting two subtree crossover points between both parents. The selected genes are swapped with the corresponding subtree in the other parent. The chromosome level mutation leads to selecting a random gene of a given parent to replace such substructure with a new randomly mutated gene.  The mutation at the gene level is calculated by applying a subtree mutation to a probabilistically selected gene; the subtree after that point is removed and replaced with a new subtree.
\subsubsection{Stop Criteria}
The evolutionary loop is terminated until one of these two conditions is reached: (1) an acceptable classification rate, or  (2) the total number of generations.

\subsection{Convolutional Neural Networks}

The ML community introduced the idea of designing DL models that build features from images. LeCun et al. \cite{lecun1989backpropagation} presented the modern framework of CNN, but the first time that CNN starts attracting attention was with the development of the AlexNet model \cite{krizhevsky2012imagenet}. The authors participated in the ImageNet Large-Scale Visual Recognition Challenge 2012, where they reduced by half the error rate on the image classification task.

AlexNet layer-architecture consists of 5 convolutional, three max-pooling, two normalization, and three fully connected layers (the last with 1000 softmax outputs), 60 Million parameters in 500,000 neurons. Additionally, Alex et al. \cite{krizhevsky2012imagenet} introduced the use of ReLU (Rectified Linear Unit) as an activation function with the benefits of much faster training than using tanh or sigmoid functions. To prevent overfitting, they also introduced the dropout and data augmentation methods.

\subsection{Adversarial Attack}

\begin{figure}
\begin{center}
\includegraphics[width=0.9\textwidth]{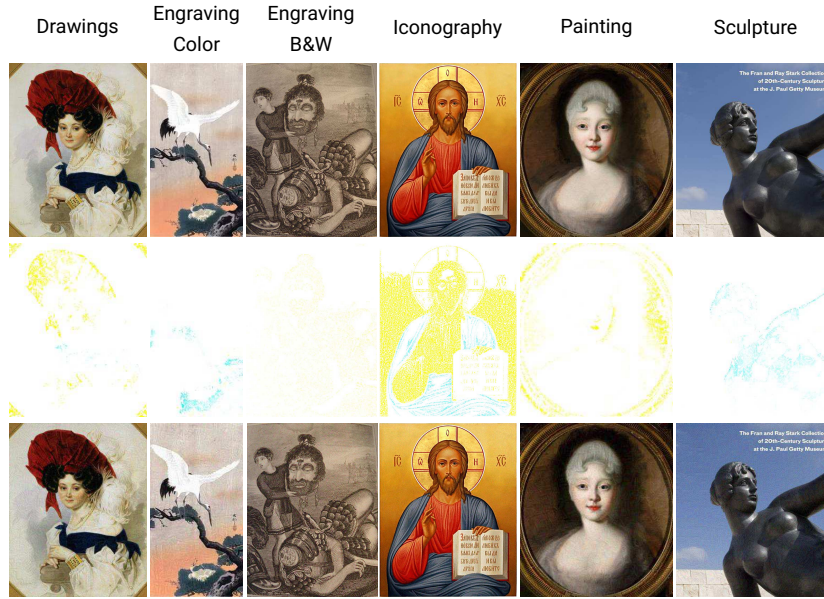}
\end{center}
\caption{Illustrations of adversarial examples from each class generated using FGSM. The first row shows an image from each class. In the second row, the perturbations computed by the FGSM are presented. The third row shows the resulting adversarial example at $\epsilon=32$, the strongest perturbation.}
\label{adversarial}
\end{figure}

The FGSM \cite{Goodfellow2015ExplainingAH} is the most popular, easy, and widely used method for computing adversarial examples from an input image, see Figure \ref{adversarial}. It increases the loss of the classifier by solving the following equation: $\rho = \epsilon\; \sign(\nabla J(\theta,\mathbf{x},y_l))$, where $\nabla J()$ computes the gradient of the cost function around the current value of the model parameters $\theta$ with the respect to the image $\mathbf{x}$ and the target label $y_l$. $\sign()$ denotes the sign function which ensures that the magnitude of the loss is maximized and $\epsilon$ is a small scalar value that restricts the norm $L_{\infty}$ of the perturbation. 

The perturbations generated by FGSM take advantage of the linearity of the deep learning models in the higher dimensional space to make the model misclassify the image. The implication of the linearity of deep learning models discovered by FSGM is that it exists transferability between models. Kurakin et al. in \cite{Kurakin2017AdversarialML} reported that after using the ImageNet database, the top-1 error rate using the perturbations generated by FGSM is around 63-69\% for $\epsilon \in \left[2, 32\right]$.

\subsection{Database Collection}

We follow the protocol and databases from the experiment of the art media categorization problem reported in \cite{ChanLey2020CategorizationOD}. The training and validation set of images are obtained from the digitized artwork database downloaded from the Kaggle website. This database comprises five categories of art media: drawing, painting, iconography, engraving, and sculpture. For class engraving, there were two different kinds of engravings. Most of them were engravings with only one color defining the art piece. The other style was Japanese engravings, which introduce color to the images. Therefore, the engraving class was split into engraving grayscale and color. For testing, a standard database WikiArt is used from which it was selected images of the same categories. Since the Wikiart engraving class is grayscale, the ukiyo-e class (Japanese engravings) from Wikiart was used as the engraving color class. Also, the set of images of the category landscapes, which are paintings from renowned artists, is added to test the painting class. Table~\ref{Tab:dataset} provides the number of images for each database.

\begin{table}[H]
\caption{Total number of images per class obtained from Kaggle and Wikiart Databases}
\begin{center}
\resizebox{\textwidth}{!}{
\begin{tabular}{p{0.15\linewidth}p{0.14\linewidth}p{0.14\linewidth}p{0.14\linewidth}p{0.14\linewidth}p{0.14\linewidth}p{0.14\linewidth}p{0.14\linewidth}}
\hline
 		& Drawings & Engraving gray scale & Engraving color & Painting       & Iconography    & Sculpture 	& Caltech Background  \\ \hline  \hline
Train 		& 553 & 426 & 30 & 1021 		& 1038	& 868 &233		\\ \hline  \hline
Validation & 553 & 284 & 19 & 1021 & 1038	& 868  &233 \\ \hline
Wikiart & 204 & 695 & 1167 & 2089	& 251	& 116 & 233 \\ \hline 
Wikiart Landscapes &&&&136	&&&\\ \hline

\end{tabular}
}
 \end{center}
\label{Tab:dataset}
\end{table}

\section{Results}\label{Discusion}

In this section, we present and discuss the experimental results summarized in Tables~\ref{tab1} and \ref{tab2}. Table~\ref{tab1} provides results for the five classes of the Kaggle database. Each method presents its performance for training, validation, and the adversarial examples from the FGSM computed with the validation database using $\epsilon={2,4,8,16,32}$. Table~\ref{tab2} shows the result for the Wikiart images where both methods were tested. It is shown the outcome of the model for the clean images as well as the adversarial examples.

We observe in Table~\ref{tab1} that AlexNet surpassed BP in almost every class when considering the validation database except for the painting class. However, as we add perturbations to the validation images, the effect of AA becomes more notable. It is shown how the performance of AlexNet deteriorates in proportion to the AA. In the worst-case--Engraving color images--there is a drop in performance from 94.72\% to 17.22\% of classification accuracy. On the other hand, BP preserves its performance on all experiments even when we added the most substantial perturbation of $\epsilon=32$. Hence, if we look at each of the comparisons (bold numbers), BP outperforms AlexNet.

For the testing part (see Table~\ref{tab2}), we have that BP obtained notable better results for painting, painting landscapes, and drawings. In contrast, AlexNet obtained superior performance on engraving grayscale, engraving color, and iconography. We should mention that in any case, the results of both methods are very good. For the sculpture class, BP matches the performance of AlexNet with a difference of around 0.6\%. Then again, the susceptibility of AlexNet to the AA is a significant problem. Its accounts fall abruptly on all classes; meanwhile, the BP output remains steady.

\begin{table}[h]
\caption{Results obtained after applying BP and AlexNet on the Kaggle database. Each method presents its classification accuracy for training, validation, and the adversarial examples using FGSM computed from the validation database at $\epsilon={2,4,8,16,32}$}
\label{tab1}
\begin{center}
\resizebox{\textwidth}{!}
{
\begin{tabular}{|l|l|l|l|l|l|l|l|l|l|l|l|l|l|l|}
\hline
 & \multicolumn{7}{l|}{Brain Programming (BP)} & \multicolumn{7}{l|}{AlexNet} \\ \hline
& train  & val & $\epsilon2$ & $\epsilon4$ & $\epsilon8$ & $\epsilon16$ & $\epsilon32$ & train  & val & $\epsilon2$ & $\epsilon4$ & $\epsilon8$ & $\epsilon16$ & $\epsilon32$  \\ \hline
Sculpture   & 93.26  & 92.79 & \textbf{92.79} & \textbf{92.79} & \textbf{92.79} & \textbf{92.79} & \textbf{92.79} & 99.36 & \textbf{95.78} & 90.93 & 90.93 & 63.24 & 27.5 & 14.57 \\ \hline
Painting    & 99.68 & \textbf{99.04} & \textbf{98.25} & \textbf{98.25} & \textbf{98.48} & \textbf{98.41} & \textbf{98.48} &  98.96  &  97.69 & 93.46 & 93.46 & 83.01 & 66.99 & 69.30 \\ \hline
Engraving gray scale  & 89.76 & 92.05 & 92.23 & 92.23 & \textbf{92.23} & \textbf{91.70} & \textbf{91.87} & 99.76 & \textbf{99.29} & \textbf{96.11} & \textbf{96.11} & 78.62 & 56.71 & 47.88  \\ \hline
Engraving color  & 98.33 & 97.37 & \textbf{97.37} & \textbf{97.37} & \textbf{97.37} & \textbf{97.37} & \textbf{97.37} & 100 & \textbf{100} & 73.68 & 73.68 & 23.68 & 13.16 & 15.79 \\ \hline
Iconography & 92.84 & 91.42 & 91.42 & 91.42 & \textbf{91.42} & \textbf{91.42} & \textbf{91.42} & 99.61 & \textbf{98.66} & \textbf{96.30} & \textbf{96.30} & 83.24 & 52.26 & 38.39 \\ \hline
Drawings    & 96.56 & 90.59 & \textbf{90.59} & \textbf{90.59} & \textbf{90.59} & \textbf{90.59} & \textbf{90.59} & 96.44 & \textbf{91.35} & 85.75 & 85.75 & 66.79 & 44.91 & 35.62 \\ 
\hline
\end{tabular}
}
\end{center}
\end{table}

\begin{table}[h]
\caption{Results obtained after applying BP and AlexNet on the Wikiart database. Each method presents its classification accuracy for testing, and the adversarial examples using FGSM computed from the test database at $\epsilon={2,4,8,16,32}$}\label{tab2}
\begin{center}
\resizebox{\textwidth}{!}{
\begin{tabular}{|l|l|l|l|l|l|l|l|l|l|l|l|l|}
\hline
 & \multicolumn{6}{l|}{Brain Programming (BP)} & \multicolumn{6}{l|}{AlexNet} \\ \hline
& test & $\epsilon2$ & $\epsilon4$ & $\epsilon8$ & $\epsilon16$ & $\epsilon32$ & test & $\epsilon2$ & $\epsilon4$ & $\epsilon8$ & $\epsilon16$ & $\epsilon32$  \\ \hline
Sculpture   & 90.54 & \textbf{90.83} & \textbf{90.83} & \textbf{90.83} & \textbf{90.83} & \textbf{90.83} & \textbf{91.15} & 87.61 & 87.61 & 65.49 & 44.25 & 36.87\\ \hline
Painting    & \textbf{100} & \textbf{95.65} & \textbf{95.65} & \textbf{95.65} & \textbf{95.65} & \textbf{95.65} & 94.06 &  90.57  &  90.57 & 64.64 & 41.04 & 41.00 \\ \hline
Painting Landscapes & \textbf{100} & \textbf{100} & \textbf{100} & \textbf{100} & \textbf{100} & \textbf{100} & 93.77 &  86.99  &  86.99 & 61.25 & 41.46 & 35.77 \\ \hline
Engraving gray scale  & 91.55 & 92.64 & 92.64 & \textbf{91.97} & \textbf{91.72} & \textbf{91.63} & \textbf{98.58} & \textbf{94.06} & \textbf{94.06} & 75.06 & 57.32 & 54.64 \\ \hline
Engraving color   & 89.92 & \textbf{89.68} & \textbf{89.68} & \textbf{89.74} & \textbf{89.86} & \textbf{89.80} & \textbf{94.72} & 73.55 & 73.55 & 25.49 & 12.30 & 17.22 \\ \hline
Iconography & 91.74 & 91.66 & 91.66 & \textbf{91.82} & \textbf{91.74} & \textbf{91.74} & \textbf{96.07} & \textbf{93.39} & \textbf{93.39} & 70.04 & 37.40 & 28.72 \\ \hline
Drawings    & \textbf{94.05} & \textbf{94.28} & \textbf{94.28} & \textbf{93.59} & \textbf{93.81} & \textbf{94.50} & 86.73 & 77.8 & 77.8 & 57.21 & 41.19 & 32.72 \\ \hline
\end{tabular}
}
 \end{center}
\end{table}

\section{Conclusions and Future work}
In conclusion, AA are a severe threat to the security of DL models. Their performance can be extremely weakened with such small perturbations. With traditional CV approaches, it is not easy to obtain results comparable to DL models. However, we propose a GP-like methodology inspired by the brain's behavior to solve art media classification. This work innovates compared with a DL model by considering performance and robustness against adversarial attacks. Also, we want to extend the robustness to adversarial attacks using CV mainstream approaches from image classification for future work. Furthermore, we will increase the number of adversarial attacks to assess the classifiers' performance under various conditions.

%
%
%
\bibliographystyle{splncs04}
\bibliography{references}
\end{document}